\documentclass[letterpaper]{article} % DO NOT CHANGE THIS
\usepackage{aaai24}  % DO NOT CHANGE THIS
\usepackage{times}  % DO NOT CHANGE THIS
\usepackage{helvet}  % DO NOT CHANGE THIS
\usepackage{courier}  % DO NOT CHANGE THIS
\usepackage[hyphens]{url}  % DO NOT CHANGE THIS
\usepackage{graphicx} % DO NOT CHANGE THIS
\usepackage{natbib}  % DO NOT CHANGE THIS AND DO NOT ADD ANY OPTIONS TO IT
\usepackage{caption} % DO NOT CHANGE THIS AND DO NOT ADD ANY OPTIONS TO IT
\usepackage{algorithm}
\usepackage{algorithmic}
\usepackage{float}

\usepackage{booktabs}
\usepackage{multirow}
\usepackage{bigstrut}
\usepackage{amsmath}
\usepackage{amssymb}
\graphicspath{{images/} }

\usepackage{newfloat}
\usepackage{listings}
\DeclareCaptionStyle{ruled}{labelfont=normalfont,labelsep=colon,strut=off} % DO NOT CHANGE THIS
\floatstyle{ruled}
\newfloat{listing}{tb}{lst}{}
\floatname{listing}{Listing}
\nocopyright % -- Your paper will not be published if you use this command
\title{FMViT: A multiple-frequency mixing Vision Transformer}
\iftrue
\author {
    Wei Tan,
    Yifeng Geng,
    Xuansong Xie
}
\affiliations {
    DAMO Academy, Alibaba Group\\
    ziyuan.tw@alibaba-inc.com, cangyu.gyf@alibaba-inc.com, xingtong.xxs@taobao.com
}
\fi

\usepackage{bibentry}
\begin{document}

\maketitle

\begin{abstract}
The transformer model has gained widespread adoption in computer vision tasks in recent times. However, due to the quadratic time and memory complexity of self-attention, which is proportional to the number of input tokens, most existing Vision Transformers (ViTs) encounter challenges in achieving efficient performance in practical industrial deployment scenarios, such as TensorRT and CoreML, where traditional CNNs excel. Although some recent attempts have been made to design CNN-Transformer hybrid architectures to tackle this problem, their overall performance has not met expectations. To tackle these challenges, we propose an efficient hybrid ViT architecture named FMViT. This approach enhances the model's expressive power by blending high-frequency features and low-frequency features with varying frequencies, enabling it to capture both local and global information effectively. Additionally, we introduce deploy-friendly mechanisms such as Convolutional Multi-group Reparameterization (gMLP), Lightweight Multi-head Self-Attention (RLMHSA), and Convolutional Fusion Block (CFB) to further improve the model's performance and reduce computational overhead. Our experiments demonstrate that FMViT surpasses existing CNNs, ViTs, and CNN-Transformer hybrid architectures in terms of latency/accuracy trade-offs for various vision tasks. On the TensorRT platform, FMViT outperforms Resnet101 by 2.5\% (83.3\% vs. 80.8\%) in top-1 accuracy on the ImageNet dataset while maintaining similar inference latency. Moreover, FMViT achieves comparable performance with EfficientNet-B5, but with a 43\% improvement in inference speed. On CoreML, FMViT outperforms MobileOne by 2.6\% in top-1 accuracy on the ImageNet dataset, with inference latency comparable to MobileOne (78.5\% vs. 75.9\%). 
Our code can be found at https://github.com/tany0699/FMViT.

\end{abstract}

\begin{figure}[h]
    \centering
    \includegraphics[width=0.45\textwidth]{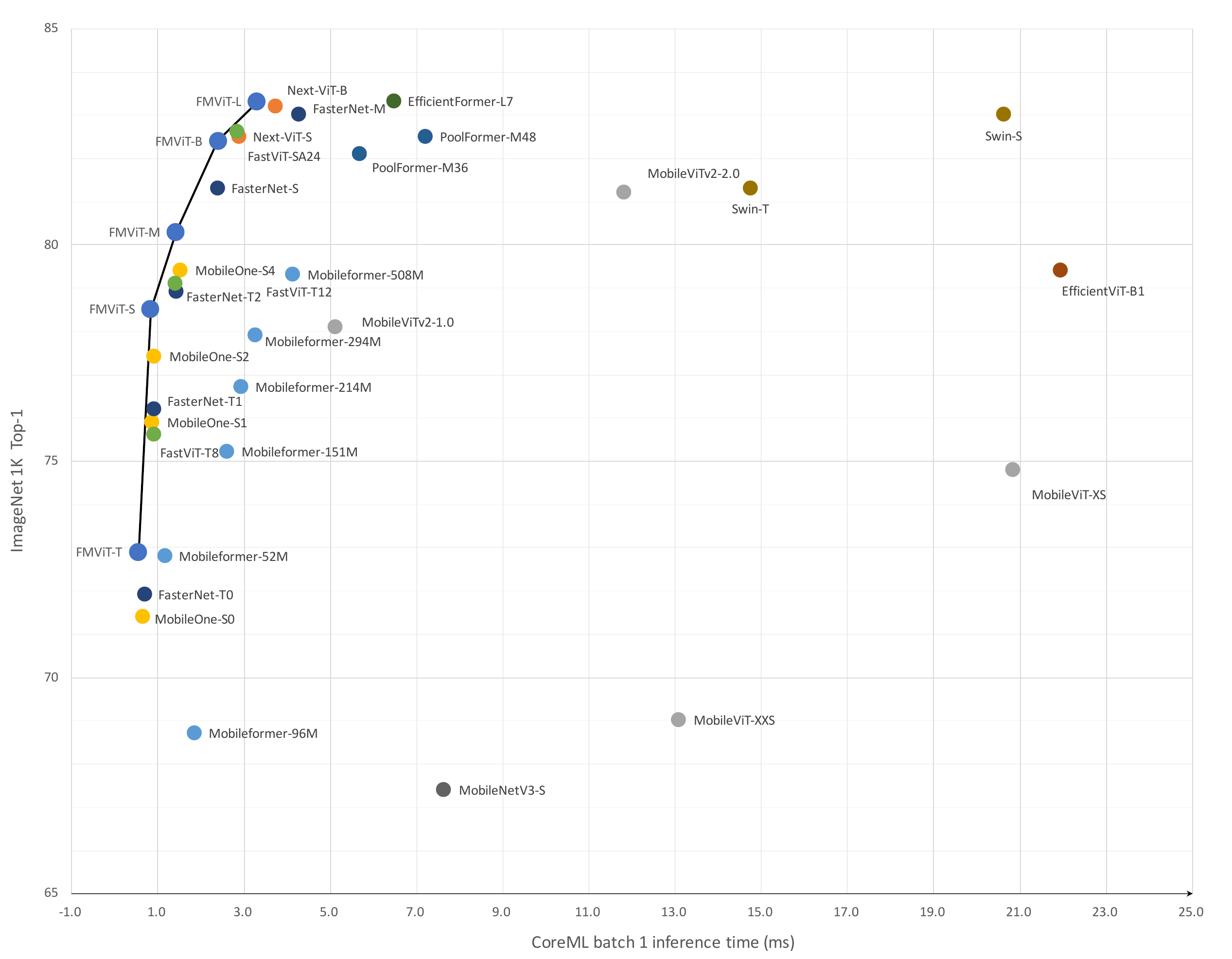}
    \caption{Speed-performance trade-off on ImageNet1K}
    \label{fig:fig0}
\end{figure}

\section{Introduction}
Vision Transformers (ViTs) have recently succeeded in various computer vision applications such as image classification, object detection, and semantic segmentation and have received extensive attention from industry and academia. Despite this, Convolutional Neural Networks (CNNs) remain the preferred choice for real-world vision tasks, primarily because ViTs typically exhibit slower performance than traditional CNNs, such as ResNets. The inference speed of Transformer models is constrained by elements such as the Multi-head Self-attention (MHSA) mechanism, non-fusible LayerNorm, and GELU layers, along with frequent memory accesses.

Numerous endeavors have been undertaken to liberate Vits from the high-latency issue. For instance, models such as Swin \cite{DBLP:conf/iccv/LiuL00W0LG21}, PoolFormer \cite{DBLP:conf/cvpr/YuLZSZWFY22}, Reformer \cite{DBLP:conf/iclr/KitaevKL20}, MaxViT \cite{DBLP:conf/eccv/TuTZYMBL22}, SepViT \cite{DBLP:journals/corr/abs-2203-15380}, and MobileViT \cite{DBLP:journals/corr/abs-2203-15380}, among others, strive to develop spatial attention methods that are more efficient and mitigate the quadratic surge in computational complexity of MHSA. Concurrently, other initiatives, including EfficientFormer \cite{DBLP:conf/nips/LiYWHETWR22} and MobileViT \cite{DBLP:journals/corr/abs-2203-15380}, are exploring ways to construct CNN-Transformer hybrid architectures that balance accuracy and latency. This is achieved by integrating effective convolutional blocks with potent Transformer blocks. Notably, most of the current state-of-the-art (SOTA) models are designed as CNN-Transformer hybrids. These models predominantly utilize convolutional blocks in the preliminary stages and reserve the stacking of Transformer blocks for the final stages.

Presently, neither the Convolutional Block nor the Transformer Block can simultaneously achieve efficiency and performance in existing works. Although the precision-latency tradeoff has improved over the years, the overall performance of modern hybrid systems still needs to improve. This study introduces four critical components for designing effective vision Transformer networks to address these challenges. Firstly, inspired by NextViT's \cite{DBLP:journals/corr/abs-2207-05501} mixing of high-frequency features and low-frequency features, a potent Multi-Frequency Fusion Block (FMB) is introduced, amalgamating multiple high-frequency and low-frequency features to enhance the model's information flow and expressive capacity. Secondly, a Lightweight Convolution Fusion Block (CFB) is proposed to efficiently blend the local modeling capability of convolution with convolution multi-group reparameterization, further bolstering modeling performance. Thirdly, convolutional multi-group reparameterization is suggested. It integrates the spatial information of different subchannels during the training phase. It fuses them into a convolution in the inference phase, improving the model's accuracy while maintaining the inference speed. Lastly, a lightweight self-attention block, termed RLMHSA, is developed. It employs a lightweight and reparameterized design to augment the modeling ability and expedite the inference stage.

A CNN-Transformer hybrid architecture, FMViT, is introduced based on the above methods. Analogous to NextViT \cite{DBLP:journals/corr/abs-2207-05501}, the use of TensorRT and CoreML signifies real deployed architectures in server-side and mobile devices, respectively, with their inference latency representing the actual time consumption in the industry. As depicted in Figure \ref{fig:fig0}, FMViT achieves an optimal balance between delay and accuracy in the ImageNet-1K classification task. On TensorRT, FMViT surpasses Resnet101 by 2.5\% in top-1 accuracy on the ImageNet dataset, maintaining a comparable inference latency. Concurrently, it exhibits performance on par with EfficientNet-B5, enhancing the inference speed by 43\%. On CoreML, the top-1 accuracy on the ImageNet dataset exceeds MobileOne by 2.6\% while maintaining a similar inference latency.

Our major contributions are outlined below:
\begin{itemize}
\item An efficient multi-frequency Fusion Block (FMB) is proposed to combine multiple sets of high-frequency and low-frequency features, enhance the information flow of the model, and enhance the expression ability of the model.

\item Proposes a lightweight Convolutional Fusion Block (CFB), which efficiently blends the local modeling capabilities of Convolutions and uses convolutional multi-group reparameterization to further provide modeling performance.

\item Convolutional multi-group reparameterization is proposed, which fuses the spatial information of different subchannels in the training stage and fuses it into a convolution in the inference stage, so as to improve the accuracy of the model while the inference speed is unchanged.

\item  Multiple groups of Multilayer Perceptron Layer (gMPL) blocks were proposed to fuse global signals and local information to enhance the expression ability of the model.

\item Proposes a Lightweight Self-Attention Block (RLMHSA), which adopts a lightweight and reparameterized design, enhances the global modeling ability of the module, and improves the speed of the inference stage.
\end{itemize}

\section{Related Work}
\subsection{Convolutional Networks}
Convolutional Neural Networks (CNNs) have been the de facto vision architecture standard for various computer vision applications, such as semantic segmentation, object identification, and image classification, since 2012. ResNet \cite{DBLP:conf/cvpr/HeZRS16} uses residual connections to stop the network from deteriorating and keep the network deep and able to capture high-level abstractions. On the other hand, DenseNet \cite{DBLP:journals/corr/HuangLW16a} promotes the concatenation of feature maps and the reuse of features. Convolution is introduced point-wise and depth-wise by MobileNets \cite{DBLP:journals/corr/HowardZCKWWAA17, DBLP:conf/cvpr/SandlerHZZC18} to create models with less memory usage and quicker response times. By using group point-wise convolution and channel shuffling, ShuffleNet \cite{DBLP:conf/cvpr/ZhangZLS18} substantially lowers computing expenses. According to ShuffleNetv2 \cite{DBLP:conf/eccv/MaZZS18}, network architecture design should prioritize direct metrics, such as speed, over indirect metrics, like FLOPs. ConvNeXt \cite{DBLP:conf/eccv/MaZZS18} explores the structure of vision Transformers and proposes a pure CNN model that, while retaining the simplicity and efficiency of traditional CNNs, can effectively compete with state-of-the-art hierarchical vision Transformers across a range of computer vision benchmarks.

\subsection{Vision Transformers}
The concept of the Transformer was initially introduced in the field of natural language processing (NLP). ViT \cite{DBLP:conf/iclr/DosovitskiyB0WZ21}, in implementing self-attention, segments the image into patches and treats these patches as words, thereby demonstrating the Transformer's efficacy in various vision-related tasks. The teacher-student method proposed by DeiT \cite{DBLP:conf/iclr/DosovitskiyB0WZ21} is designed explicitly for Transformers. T2T-ViT \cite{DBLP:conf/iccv/0007CWYSJTFY21} introduces a unique token-to-token process to progressively tokenize images into tokens and aggregate them structurally. The Swin Transformer \cite{DBLP:conf/iccv/LiuL00W0LG21} introduces a universal backbone that constructs hierarchical features with a computational cost linearly proportional to the image size. Meanwhile, PiT \cite{DBLP:conf/iccv/HeoYHCCO21} incorporates a pooling layer into ViT and conducts experiments to validate its efficacy.

\subsection{Hybrid Models}
Recent studies indicate that a hybrid design \cite{DBLP:conf/icml/ZhouYXXAFA22, DBLP:conf/nips/LiYWHETWR22, DBLP:conf/cvpr/SrinivasLPSAV21, DBLP:conf/iclr/MehtaR22}, integrating both convolution and Transformer, effectively leverages the strengths of both architectures. BoTNet \cite{DBLP:conf/cvpr/SrinivasLPSAV21} employs global self-attention to supplant the spatial convolutions of the final three bottleneck blocks in ResNet. Concurrently, lightweight and efficient ViTs, such as MobileViT \cite{DBLP:conf/iclr/MehtaR22} and MobileViTv2 \cite{DBLP:conf/iclr/MehtaR22}, have been introduced for mobile devices. The fusion of Mobile-Former  \cite{DBLP:conf/cvpr/ChenDCLDY022} with the proposed lightweight cross-attention model enhances computational efficiency and boosts representational power. EfficientFormer \cite{DBLP:conf/nips/LiYWHETWR22} and EfficientFormerV2 \cite{DBLP:journals/corr/abs-2212-08059} adhere to size-consistent designs, seamlessly employing hardware-friendly 4D MetaBlocks and potent 3D MHSA blocks for joint size and speed search via NAS. ToMe presents a ViT model that accelerates without training. BiFormer establishes an efficient pyramid network architecture through bidirectional attention routing. NextViT \cite{DBLP:journals/corr/abs-2207-05501} captures one high-frequency feature and one low-frequency feature in the network separately, which are then blended to enhance the modeling capabilities of the model.

\subsection{Structural Reparameterization}
Reparameterization employs complex modules to enhance model performance during the training phase. It consolidates these complex modules into simpler ones during the inference phase, following the linear principle of the convolution operator. This process aims to boost the model's inference speed without compromising performance. ACNet \cite{DBLP:conf/iccv/DingGDH19} pioneered reparameterization to merge 3x3 convolutions into a 1x1 convolution, while RepVGG  \cite{DBLP:conf/cvpr/Ding0MHD021} applied reparameterization to skip connections, thereby reducing memory access costs. DBB \cite{DBLP:conf/cvpr/Ding0HD21} further expanded upon six prevalent reparameterization methods. The concept of linear training time overparameterization was introduced to augment the power of such models \cite{DBLP:conf/cvpr/0020YZD00022, DBLP:journals/corr/abs-2204-00826}. MobileOne  \cite{DBLP:journals/corr/abs-2206-04040} employs over-parameterization to enhance the performance of vision converters for mobile devices.

\section{Methods}
This section introduces the proposed FMViT architecture, followed by a discussion and analysis of its key designs. These include the Convolutional Fusion Module (CFB), the Multi-frequency Mixing Module (FMB), the Lightweight Multi-head Attention Module (RLMHSA), the Convolutional Multi-group Reparameterization method (gMLP), and the MLP module constructed using this method.

\begin{figure*}[h]
    \centering
    \includegraphics[width=1.0\textwidth]{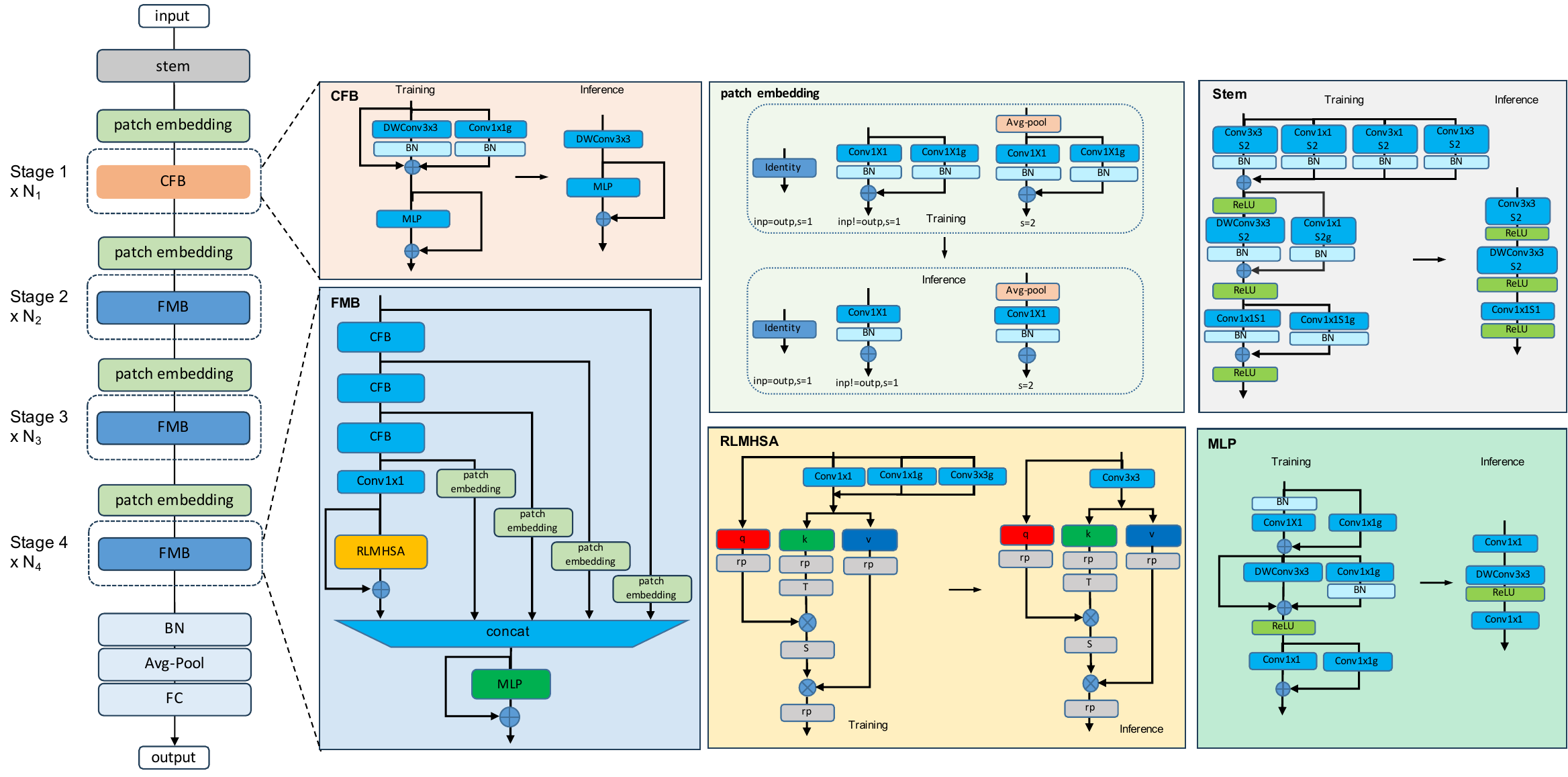}
    \caption{The figure shows the overall structure of FMViT. It mainly includes the Convolutional Fusion Block (CFB), the multi-frequency Fusion Block (FMB), the lightweight Multi-head Attention Module (RLMHSA), and the parameterized Multi-layer Perceptron Module (gMLP).}
    \label{fig:fig1}
\end{figure*}

\begin{figure*}[h]
    \centering
    \includegraphics[width=0.7\textwidth]{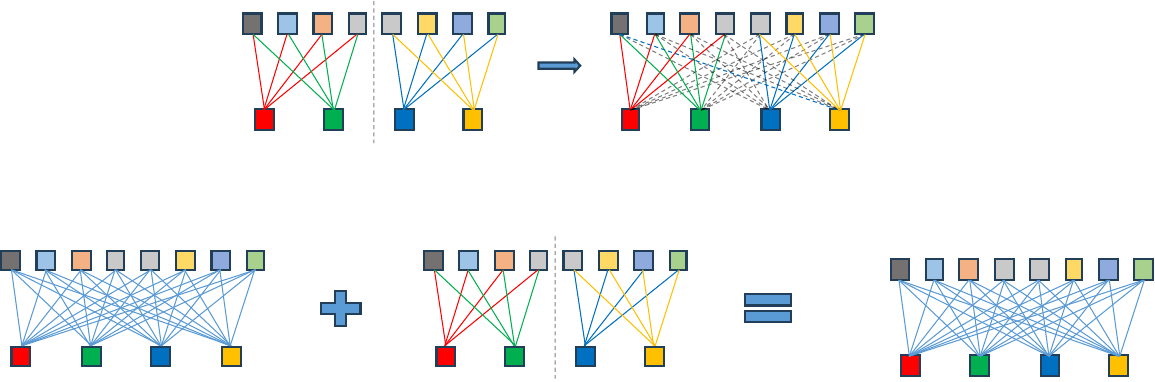}
    \caption{Top: A schematic depiction illustrating the notion of convolutional multi-group reparameterization, CONV($K_{hi}$, $K_{wi}$, $G_i$) is comparable to sparse CONV($K_{h}$, $K_{w}$, G). Bottom: By reparameterizing numerous groups of convolutions in the training phase, different groups of convolutions in the training phase are equal to a single convolution in the inference phase.}
    \label{fig:fig2}
\end{figure*}

\subsection{Overview}
Figure \ref{fig:fig1} illustrates the comprehensive FMViT architecture. FMViT utilizes a conventional pyramidal structure, with each stage comprising a downsampling module and a Convolutional Fusion Module (CFB) with convolution only or a Multi-frequency Mixing Module (FMB) with transformers. The stem reduces the spatial resolution to a quarter of the original input image, and each subsequent stage progressively halves the spatial resolution while incrementally increasing the number of channels. We explore the information interaction module and, inspired by MetaFormer \cite{DBLP:conf/cvpr/YuLZSZWFY22}, introduce the Convolutional Fusion Module (CFB) to address short-range data dependencies. A Multi-frequency Mixing Module (FMB) is proposed to further fuse local and global data by decomposing multiple frequency bands. This multi-channel frequency fusion enhances the model's capacity for modeling. To decrease the computational complexity of Multi-head Attention (MHSA), we propose a lightweight RLMHSA module. The model's inference speed is improved without significantly compromising accuracy through parameter sharing and re-parameterization. These core modules develop a series of CNN-Transformer hybrid architectures that achieve an optimal balance between accuracy and latency on mobile-side CPUs and server-side GPUs, surpassing the state-of-the-art model.

\subsection{Convolutional multi-group reparameterization}
The 1x1 convolution is a linear fusion or channel conversion mechanism with global modeling capacity. The translation invariance characteristic of the kxk convolution is utilized to depict local space. The lack of local modeling between specific neighboring channel features limits ungrouped convolution operators' efficient information fusion capabilities. During the training phase, we suggest group reparameterization of kxk convolutions, initially defining the convolution operation as CONV($K_h$,$K_w$,G), where $K_h$ and $K_w$ represent the convolution kernel sizes, and G denotes the convolution group size.

Assuming that the original convolution is defined as CONVA=CONV($K_h$,$K_w$,G), during the training phase, multiple convolutions with diverse group sizes are connected in parallel:
\begin{equation}
CONVB = CONVA + \sum_{i=1}^{N}CONV(K_{hi},K_{wi},G_i)
\end{equation}
Where $\forall i \in N, G_i>=G, K_{hi}<=K_h, K_{wi}<=K_w$, and N is a predefined constant.

Figure \ref{fig:fig2} illustrates the reparameterization of CONVB into CONVA during the inference phase. Any CONV($K_{hi}$,$K_{wi}$,$G_i$) convolution is equivalent to the sparse CONV($K_{h}$,$K_{w}$, G) convolution, signifying that the weight of the dotted line in the figure remains at a constant zero, while the other weights are unchanged. Based on additivity \cite{DBLP:conf/cvpr/Ding0HD21}, the inference stage for two convolutions with the same number of groups can be reparameterized as the convolution CONVA, as depicted in the lower portion of Figure \ref{fig:fig2}. Here, the left side represents the training stage, and the right side represents the inference stage, both of which are equivalent. Convolutional multi-group reparameterization enhances model performance without affecting the primary model's inference time.

RepMLP \cite{DBLP:journals/corr/abs-2105-01883} proposes the reparameterization of the multilayer perceptron (MLP). RepMLP employs convKxK for fusion into FC, but its weight conversion is sparse, and the parameters post-conversion become KxK times the original, rendering it unsuitable for lightweight scenarios. In the Transformer, MLPs significantly contribute to performance due to their global modeling capabilities. However, the absence of local modeling capability restricts its potential. To aid the original convolutions in grouping modeling of local channels, multiple parallel conv1x1 convolutions with $G'>1$ are introduced to the two original CONV(1,1,1) convolutions, reconstructing the MLP module. This approach concurrently focuses on information from different representation subspaces at various locations for efficient local representation learning.
\begin{small}
\begin{equation}
CONV^{'}X(1,1,1) = CONVX(1,1,1) + \sum_{i=1}^{N}CONV(1,1,G_i)
\end{equation}
\end{small}
In this context, X=1 or X=2 signifies either CONV1 or CONV2. CONV1 and CONV2 represent the two pre-reparameterized convolutions of the MLP, while the post-reparameterized convolutions of the MLP are also denoted as $CONV^{'}1$ and $CONV^{'}2$.

To enhance the hybrid modeling of global and local aspects in MLP, a depthwise convolution is incorporated between the two conv1x1s. A shortcut connection ensures that the global and local information flows do not interfere, and the added depthwise convolution is reparameterized. Experimental results indicate that augmenting the depthwise convolution capacity for local information fusion enhances MLP performance by 1.96\% on Imagenet1k.

\subsection{Convolutional Fusion Block(CFB)}
Transformer blocks have demonstrated significant results across various vision tasks, with the attention-based token mixer module and the MetaFormer \cite{DBLP:conf/cvpr/YuLZSZWFY22} paradigm underscoring their inherent advantages. However, the inference speed of Transformer blocks could be more efficient, particularly on mobile devices where the performance of multi-head attention, LayerNorm, and GELU calculations is suboptimal. We have demonstrated the success of the MetaFormer paradigm and, building upon it, propose an efficient CFB module that exclusively employs depthwise separable convolutions (DWConv) as the token mixer. CFB maintains the deployment advantage of the BottleNeck block while achieving comparable performance to a Transformer block, as depicted in Figure \ref{fig:fig1}. The CFB is constructed using DWConv and MLP, adhering to the general MetaFormer design. CFB ensures performance while significantly improving inference deployment performance. Additionally, reparameterization is implemented during training to enhance further CFB's performance, where DWConv utilizes a standard, widely-used reparameterization. The MLP employs the convolutional multigroup reparameterization proposed in this work.

\subsection{Multi-frequency Fusion Block(FMB)}
While CFB has effectively learned local representation, the pressing need to handle global information collection remains. Transformer blocks capture low-frequency signals, providing global information such as shape and structure. Existing research indicates that Transformer blocks may partially diminish high-frequency information, including local texture details. To extract more fundamental and unique features, signals from various frequency bands must be meticulously integrated, as they are crucial to the human visual system.

High-frequency signals provide local information, which is integral to preserving the completeness of the information. Distinct frequency features contribute unique information, rendering high-frequency signals vulnerable to degradation by the Transformer Block. The amalgamation of various high-frequency characteristics with low-frequency features could enhance the model's information flow and expressive capacity, drawing inspiration from information distillation and frequency mixing in image super-resolution \cite{DBLP:conf/mm/HuiGYW19}. As illustrated in Figure \ref{fig:fig1}, the CFB module initially captures high-frequency features, subsequently outputting three sets of high-frequency features at different frequencies. Patch embedding fusion is then employed to splice the output of the lightweight multi-head attention module, creating a signal with both high and low frequencies. Through MLP layers, more fundamental and pronounced traits are extracted. The following formula can be expressed as follows:
\begin{align*}
z_1 &= f_1(x^{l-1}) \\
z_2 &= f_2(z_1) \\
z_3 &= f_3(z_2) \\
z_4 &= f_4(z_3) \\
z &= CONCAT(x^{l-1}, z_1, z_2, z_3, z_4) \\
x^l &= z + MLP(z)
\end{align*}
Here, $x^{l-1}$ is defined as the input of the ($l-1$)th block, while $x^l$ signifies the output of the lth block. CONCAT refers to the CAT join operation. $f_1$–$f_3$ represent high-pass filters that generate different high-frequency signals, as exemplified by CFB. $f_4$ is the low-pass filter that produces the low-frequency signal, as demonstrated by RLMHSA. 

Unlike LN and GELU, FMB consistently employs BN and ReLU as the effective norm and activation layers \cite{DBLP:journals/corr/abs-2207-05501}. These operators can be efficiently computed, especially on mobile devices, with minimal performance loss due to their speed-friendly nature.

FMB can gather and integrate multi-frequency information within a lightweight framework, thereby significantly enhancing the model's performance compared to the traditional pure Transformer module.

\subsection{Lightweight Multi-head Self-Attention (RLMHSA)}
The computational demand of the Transformer is proportional to the square of the input token dimension, making it computationally intensive when dealing with large input dimensions. Despite its relatively small number of parameters, the inference time on mobile devices, for instance, is extensive, necessitating a more lightweight design for the self-attention module. ViT-LSLA \cite{DBLP:journals/corr/abs-2210-17115} substitutes the Key (K) and Value (V) of self-attention with the original input (X) to achieve a lightweight self-attention structure. As depicted in Figure \ref{fig:fig1}, we propose a lightweight multi-head self-attention method that shares parameters and then applies reparameterization in this study. 

The original MSA is defined as follows:
\begin{equation} \label{eq3}
Atten(Q,K,V)=Softmax(QK^T)V
\end{equation}

Where $Q=XW_q$, $K=XW_k$, and $V=XW_v$, input $X \in R^{M \times d}$, Query, Key, and Value matrices $W_q$, $W_k$, and $W_v$ $\in R^{d \times d}$, respectively, M is the number of tokens, and d is the dimension of tokens. By deforming Equation \ref{eq3}, we obtain:
\begin{align*}
Atten(Q,K,V) &= Softmax(XW_q(XW_k)^T)XW_v \\
&= Softmax(XW_qW_k^TX^T)XW_v \\
&= Softmax(XW_{qk}X^T)XW_v \\
&= Softmax(X(XW_{qk}^T)^T)XWv \\
&= Atten(X,K',V) 
\end{align*}

The projection matrices of Q and K are consolidated for parameter sharing, which equates to a new matrix $W_{qk}$ with $K=XW_{qk}^T$. Moreover, allowing $W_{qk}$ and $W_v$ to share parameters, they share a projection convolution, denoted as $W = W_{qk}^T = W_v$, then:
\begin{equation}
Atten(X,K',V') = Softmax(X(XW)^T)XW
\end{equation}
Where $K'=XW, V'= XW$, and the structure is shown in Figure \ref{fig:fig1} for RLMHSA.
Consequently, a single convolution is required to map the input vector, with the K and Q vectors sharing the same convolution, thereby eliminating the need for two separate convolutions. During training, convolutional multi-group parameterization is employed to emulate the blend of local and global information that characterizes the RLMHSA module and enhances MHSA performance. Experimental results indicate that, compared to MHSA on the Imagenet1k classification task, RLMHSA reduces the parameter count by 8M, accelerates speed by 3\%, and enhances performance by 0.5\%.

\subsection{Stem block and Patch Embedding block}
The model's initial stage employs a stem with two down-sampling operations to reduce computational load, as suggested by FastViT \cite{DBLP:journals/corr/abs-2303-14189}. A lightweight structure is achieved through the use of a Conv3x3+DWConv3x3 design. Initial convolutions utilize Conv3x1 and Conv1x3 to reparameterize vertical and horizontal edge extraction.

Several scenarios exist for patch embedding: no operation is necessary if the input and output channel numbers and token dimensions are identical. Conv1x1 is employed for channel number conversion when the input and output channel numbers differ, but the token dimension remains the same. If the token dimensions vary, a lightweight downsampling operation, avg-pool, is utilized for downsampling, followed by a Conv1x1 convolution for fusion or transformation. During the training phase, convolutional multi-group reparameterization is also applied to enhance accuracy.

\subsection{FMViT Architectures}
As delineated in Table \ref{tab:table0}, this study presents five model structures for reference based on the number of parameters and model size, specifically FMViT-T, FMViT-S, FMViT-M, FMViT-B, and FMViT-L. Here, ``Channels" refers to the number of output channels from the internal submodule of the module; ``FM-Channels" denotes the number of FMB intermediate frequency division channels, and ``S" represents the stride in convolution, or Avg-pool. The expansion ratio for each MLP layer is set at 2, and the head dimension in RLMHSA is fixed at 32. In alignment with Nextvit \cite{DBLP:journals/corr/abs-2207-05501}, BatchNorm, and ReLU are employed for normalization and activation functions.

% Table generated by Excel2LaTeX from sheet 'architectures'
\begin{table}[htbp]
  \centering
  \caption{Architecture details of FMViT variants.}
  \scalebox{0.45}{
    \begin{tabular}{c|c|c|c|c|c|c|cc}
    \hline
    Stages & Output size & \multicolumn{2}{c|}{Layers} & FMViT-T & FMViT-S & FMViT-M & \multicolumn{1}{c|}{FMViT-B} & FMViT-L \bigstrut\\
    \hline
    \multirow{4}[8]{*}{Stem} & \multirow{4}[8]{*}{(H/4, W/4)} & \multicolumn{2}{c|}{\multirow{4}[8]{*}{CNN Layers}} & \multicolumn{5}{c}{Conv3x3, S=2} \bigstrut\\
\cline{5-9}          &       & \multicolumn{2}{c|}{} & \multicolumn{5}{c}{DWConv3x3, S=2} \bigstrut\\
\cline{5-9}          &       & \multicolumn{2}{c|}{} & \multicolumn{5}{c}{Conv1x1, S=1} \bigstrut\\
\cline{5-9}          &       & \multicolumn{2}{c|}{} & (32,32,32) & (48,48,48) & \multicolumn{3}{c}{(64,64,64)} \bigstrut\\
    \hline
    \multirow{4}[8]{*}{Stage1} & \multirow{4}[8]{*}{(H/4, W/4)} & \multicolumn{2}{c|}{\multirow{2}[4]{*}{Patch Embedding}} & \multicolumn{5}{c}{Conv1x1, S=1} \bigstrut\\
\cline{5-9}          &       & \multicolumn{2}{c|}{} & 32    & 48    & \multicolumn{3}{c}{64} \bigstrut\\
\cline{3-9}          &       & \multirow{2}[4]{*}{CFB} & Channels & (32,32,32) & (48,48,48) & \multicolumn{3}{c}{(64,96,96)} \bigstrut\\
\cline{4-9}          &       &       & Blocks & 3     & 3     & 3     & \multicolumn{1}{c|}{3} & 6 \bigstrut\\
    \hline
    \multirow{6}[12]{*}{Stage2} & \multirow{6}[12]{*}{(H/8, W/8)} & \multicolumn{2}{c|}{\multirow{3}[6]{*}{Patch Embedding}} & \multicolumn{5}{c}{Avg-Pool,S=2} \bigstrut\\
\cline{5-9}          &       & \multicolumn{2}{c|}{} & \multicolumn{5}{c}{Conv1x1,S=1} \bigstrut\\
\cline{5-9}          &       & \multicolumn{2}{c|}{} & 32    & 48    & \multicolumn{3}{c}{96} \bigstrut\\
\cline{3-9}          &       & \multirow{3}[6]{*}{FMB} & Channels & (32,64,80) & (48,96,160) & (96,128,160) & \multicolumn{2}{c}{(96,256,320)} \bigstrut\\
\cline{4-9}          &       &       & FM-Channels & 16    & 32    & 32    & \multicolumn{2}{c}{64} \bigstrut\\
\cline{4-9}          &       &       & Blocks & 1     & 1     & 1     & \multicolumn{2}{c}{1} \bigstrut\\
    \hline
    \multirow{6}[12]{*}{Stage3} & \multirow{6}[12]{*}{(H/16, W/16)} & \multicolumn{2}{c|}{\multirow{3}[6]{*}{Patch Embedding}} & \multicolumn{5}{c}{Avg-Pool,S=2} \bigstrut\\
\cline{5-9}          &       & \multicolumn{2}{c|}{} & \multicolumn{5}{c}{Conv1x1,S=1} \bigstrut\\
\cline{5-9}          &       & \multicolumn{2}{c|}{} & 80    & 160   & 160   & \multicolumn{2}{c}{320} \bigstrut\\
\cline{3-9}          &       & \multirow{3}[6]{*}{FMB} & Channels & (80,128,160) & (160,192,320) & (160,320,480) & \multicolumn{2}{c}{(320,384,480)} \bigstrut\\
\cline{4-9}          &       &       & FM-Channels & 32    & 64    & 96    & \multicolumn{2}{c}{96} \bigstrut\\
\cline{4-9}          &       &       & Blocks & 1     & 1     & 1     & \multicolumn{1}{c|}{2} & 5 \bigstrut\\
    \hline
    \multirow{6}[12]{*}{Stage4} & \multirow{6}[12]{*}{(H/32, W/32)} & \multicolumn{2}{c|}{\multirow{3}[6]{*}{Patch Embedding}} & \multicolumn{5}{c}{Avg-Pool,S=2} \bigstrut\\
\cline{5-9}          &       & \multicolumn{2}{c|}{} & \multicolumn{5}{c}{Conv1x1,S=1} \bigstrut\\
\cline{5-9}          &       & \multicolumn{2}{c|}{} & 160   & 320   & \multicolumn{3}{c}{480} \bigstrut\\
\cline{3-9}          &       & \multirow{3}[6]{*}{FMB} & Channels & (160,192,320) & (320,384,640) & (480,512,960) & \multicolumn{2}{c}{(480,640,1280)} \bigstrut\\
\cline{4-9}          &       &       & FM-Channels & 64    & 128   & 192   & \multicolumn{2}{c}{256} \bigstrut\\
\cline{4-9}          &       &       & Blocks & 1     & 1     & 1     & \multicolumn{2}{c}{1} \bigstrut\\
    \hline
    \end{tabular}%
    }
  \label{tab:table0}%
\end{table}%

\section{Experimental Results}
In this experiment segment, we utilize PyTorch version 1.12.1 for PyTorch inference latency and the TensorRT-8.5.3 framework for TensorRT (TRT) inference latency. Both are measured in a hardware environment with an A10 GPU and CUDA 11.3. CoreML inference latency is gauged using an iPhone 13 with iOS 16.6. All batch sizes are uniformly set to 1.

% Table generated by Excel2LaTeX from sheet 'cls'
\begin{table}[htbp]
  \centering
  \caption{Comparison of different state-of-the-art classification methods for ImageNet-1K.}
    %\resizebox{\textwidth}{100mm}{
    \scalebox{0.52}{ % 0.9 0.6
    \begin{tabular}{l|c|cc|ccc|c}
    \hline
    \multicolumn{1}{c|}{\multirow{2}[4]{*}{Method}} & Image & Param & FLOPs &       & Latency(ms) &       & Top-1 \bigstrut\\
\cline{5-7}          & Size  & (M)   & (G)   & pytorch & TRT   & CoreML & (\%) \bigstrut\\
    \hline
    MobileViT-XXS & 224   & 1.3   & 0.3   & 8.1   & 1.2   & 13.10  & 69.0  \bigstrut[t]\\
    Mobileformer-52M & 224   & 3.5   & 0.1   & 13.3  & 2.9   & 1.17  & 72.8  \\
    Mobileformer-96M & 224   & 4.6   & 0.1   & 14.3  & 3.1   & 1.86  & 68.7  \\
    RepVGG-A0 & 224   & 8.3   & 1.4   & 2.1   & 0.7   & 1.06  & 72.4  \\
    MobileNetV1 & 224   & 4.2   & 0.6   & 2.2   & 0.5   & 0.84  & 70.6  \\
    MobileNetV3-S & 224   & 2.5   & 0.1   & 5.0   & 0.7   & 7.65  & 67.4  \\
    FasterNet-T0 & 224   & 3.9   & 0.3   & 5.1   & 0.8   & 0.72  & 71.9  \\
    MobileOne-S0 & 224   & 2.1   & 0.3   & 2.5   & 0.4   & 0.65  & 71.4  \\
    MobileNetV2×1.0 & 224   & 3.5   & 0.3   & 5.0   & 0.7   & 0.87  & 71.8  \\
    DeiT-T & 224   & 5.9   & 1.2   & 5.1   & 1.2   & 1.68  & 72.2  \\
    \textbf{FMViT-T} & \textbf{224} & \textbf{2.0 } & \textbf{0.3 } & \textbf{6.7 } & \textbf{0.8 } & \textbf{0.55 } & \textbf{72.9 } \bigstrut[b]\\
    \hline
    RepVGG-B1 & 224   & 51.8  & 11.8  & 3.1   & 2.5   & 4.12  & 78.4  \bigstrut[t]\\
    RepVGG-A2 & 224   & 25.5  & 5.1   & 2.4   & 1.4   & 2.11  & 76.5  \\
    Mobileformer-151M & 224   & 7.6   & 0.2   & 18.9  & 4.3   & 2.61  & 75.2  \\
    Mobileformer-214M & 224   & 9.4   & 0.2   & 20.1  & 4.6   & 2.93  & 76.7  \\
    Mobileformer-294M & 224   & 11.4  & 0.3   & 19.5  & 5.0   & 3.27  & 77.9  \\
    FasterNet-T1 & 224   & 7.6   & 0.9   & 5.4   & 0.9   & 0.93  & 76.2  \\
    MobileOne-S1 & 224   & 4.8   & 0.8   & 2.4   & 0.6   & 0.87  & 75.9  \\
    MobileOne-S2 & 224   & 7.8   & 1.3   & 2.5   & 0.6   & 0.92  & 77.4  \\
    EfficientNet-B0 & 224   & 5.3   & 0.4   & 7.9   & 1.1   & 1.64  & 77.1  \\
    FastViT-T8 & 256   & 3.6   & 0.7   & 4.9   & 1.3   & 0.92  & 75.6  \\
    EfficientFormerV2-S0 & 224   & 3.5   & 0.4   & 10.2  & 1.3   & 0.89  & 75.7  \\
    MobileViTv2-1.0 & 256   & 4.9   & 1.8   & 8.3   & 2.4   & 5.12  & 78.1  \\
    MobileViT-XS & 224   & 2.3   & 0.8   & 8.6   & 1.4   & 20.84  & 74.8  \\
    \textbf{FMViT-S} & \textbf{224} & \textbf{6.4 } & \textbf{0.8 } & \textbf{7.0 } & \textbf{1.1 } & \textbf{0.83 } & \textbf{78.5 } \bigstrut[b]\\
    \hline
    FasterNet-T2 & 224   & 15.0  & 1.9   & 5.1   & 1.2   & 1.44  & 78.9  \bigstrut[t]\\
    Mobileformer-508M & 224   & 14.0  & 0.5   & 19.9  & 5.7   & 4.14  & 79.3  \\
    EfficientNet-B1 & 224   & 7.8   & 0.6   & 11.4  & 1.6   & 2.08  & 79.1  \\
    EfficientViT-B1 & 224   & 9.1   & 0.5   & 8.8   & 0.7   & 21.95  & 79.4  \\
    FastViT-T12 & 256   & 6.8   & 1.4   & 6.0   & 1.7   & 1.42  & 79.1  \\
    MobileOne-S4 & 224   & 14.8  & 3.0   & 4.5   & 1.2   & 1.52  & 79.4  \\
    DeiT-S & 224   & 22.0  & 4.6   & 5.2   & 2.0   & 3.74  & 79.8  \\
    PoolFormer-S24 & 224   & 21.1  & 3.4   & 9.8   & 3.9   & 2.45  & 80.3  \\
    ResNeXt101-32x4d & 224   & 44.2  & 8.0   & 13.5  & 3.9   & 3.65  & 78.8  \\
    EfficientFormer-L1  & 224   & 12.3  & 1.3   & 6.9   & 1.3   & 1.57  & 79.2  \\
    \textbf{FMViT-M} & \textbf{224} & \textbf{12.8 } & \textbf{2.0 } & \textbf{7.1 } & \textbf{1.5 } & \textbf{1.42 } & \textbf{80.3 } \bigstrut[b]\\
    \hline
    EfficientNet-B3 & 224   & 12.0  & 1.0   & 12.5  & 2.0   & 2.71  & 81.6  \bigstrut[t]\\
    MobileViTv2-2.0 & 256   & 18.5  & 7.2   & 8.2   & 3.7   & 11.83  & 81.2  \\
    EfficientViT-B2 & 224   & 24.3  & 1.6   & 12.0  & 2.4   & 36.90  & 82.1  \\
    FasterNet-S & 224   & 31.1  & 4.5   & 6.3   & 2.2   & 2.41  & 81.3  \\
    ResNeSt50 & 224   & 27.5  & 5.4   & 13.0  & 2.9   & 32.90  & 81.1  \\
    ConvNeXt-T & 224   & 29.0  & 4.5   & 5.6   & 3.5   & 68.00  & 82.1  \\
    Swin-T & 224   & 29.0  & 4.5   & 8.0   & 2.4   & 14.75  & 81.3  \\
    PoolFormer-S36 & 224   & 31.2  & 5.0   & 13.0  & 5.7   & 3.40  & 81.4  \\
    EfficientFormer-L3  & 224   & 31.4  & 3.9   & 10.3  & 2.6   & 2.61  & 82.4  \\
    ResNet101 & 224   & 44.6  & 7.9   & 12.7  & 3.4   & 3.46  & 80.8  \\
    RegNetY-8G & 224   & 39.2  & 8.0   & 11.7  & 3.5   & 3.65  & 81.7  \\
    ResNet152 & 224   & 60.2  & 4.0   & 19.0  & 6.1   & 4.54  & 81.7  \\
    PoolFormer-M36 & 224   & 56.1  & 8.8   & 13.4  & 7.0   & 5.68  & 82.1  \\
    FastViT-SA24 & 256   & 20.6  & 3.8   & 10.1  & 3.4   & 2.84  & 82.6  \\
    Next-ViT-S & 224   & 31.7  & 5.8   & 13.5  & 2.8   & 2.90  & 82.5  \\
    \textbf{FMViT-B} & \textbf{224} & \textbf{24.3 } & \textbf{4.2 } & \textbf{9.1 } & \textbf{2.4 } & \textbf{2.40 } & \textbf{82.4 } \bigstrut[b]\\
    \hline
    EfficientNet-B5 & 224   & 30.0  & 2.4   & 20.1  & 4.5   & 4.66  & 83.6  \bigstrut[t]\\
    PoolFormer-M48 & 224   & 73.0  & 11.6  & 17.4  & 9.2   & 7.21  & 82.5  \\
    RegNetY-16G & 224   & 83.6  & 16.0  & 13.4  & 5.9   & 6.67  & 82.9  \\
    EfficientFormer-L7  & 224   & 82.2  & 10.2  & 15.4  & 5.1   & 6.49  & 83.3  \\
    FasterNet-M & 224   & 53.5  & 8.7   & 9.4   & 3.7   & 4.28  & 83.0  \\
    ResNeSt101 & 224   & 48.0  & 10.2  & 25.0  & 5.7   & 42.20  & 83.0  \\
    ConvNeXt-S & 224   & 50.0  & 8.7   & 10.1  & 6.5   & 147.30  & 83.1  \\
    Swin-S & 224   & 50.0  & 8.7   & 15.3  & 4.3   & 20.63  & 83.0  \\
    Next-ViT-B & 224   & 44.8  & 8.3   & 20.5  & 4.0   & 3.75  & 83.2  \\
    \textbf{FMViT-L} & \textbf{224} & \textbf{35.3 } & \textbf{7.1 } & \textbf{15.2 } & \textbf{3.9 } & \textbf{3.30 } & \textbf{83.3 } \bigstrut[b]\\
    \hline
    \end{tabular}%
    }
  \label{tab:table1}%
\end{table}%

\subsection{ImageNet-1K Classification}
\subsubsection{Implementation}
We executed an image classification experiment on the ImageNet-1K \cite{DBLP:journals/ijcv/RussakovskyDSKS15} dataset, comprising approximately 1.28 million training images and 50,000 validation images across 1,000 categories. To maintain fairness, we replicated the training parameters of the most recent vision Transformer with minor adjustments. All FMViT variants underwent training on eight V100 GPUs with a total batch size 2048 for 300 iterations. The input image was resized to a resolution of 224 x 224. Utilizing a weight decay of 0.1, we employed the AdamW \cite{DBLP:conf/iclr/LoshchilovH19} optimizer. For all FMViT variants, the learning rate was gradually reduced based on the cosine strategy, starting at 4e-5, and a linear warm-up approach was used for 20 epochs.

\subsubsection{Comparison with State-of-the-art Models}
As illustrated in Table \ref{tab:table1}, our method achieves an optimal balance of accuracy and latency when juxtaposed with recent state-of-the-art techniques such as CNNs, ViTs, and hybrid networks. When benchmarked against renowned CNNs like ResNet101 \cite{DBLP:conf/cvpr/HeZRS16}, FMViT surpasses ResNet101 by 2.5\% in top-1 accuracy on the ImageNet dataset (83.3\% vs. 80.8\%) and is 45\% faster on CoreML (3.5 ms vs. 2.4 ms). Concurrently, its performance is on par with EfficientNet-B5 \cite{DBLP:conf/icml/TanL19}, with a 43\% improvement in inference speed. In the context of ViT, FMViT-B outperforms Swin-T \cite{DBLP:conf/iccv/LiuL00W0LG21} by being up to 6x faster on CoreML and TensorRT, yet with a 1.1\% superior performance. FMViT-S exceeds DeiT-T \cite{DBLP:conf/icml/TouvronCDMSJ21} by 6.3\% on TensorRT at an equivalent inference speed (78.5\% vs. 72.2\%). FMViT-S surpasses CoreML by 8.1 percent (80.3\% vs. 72.2\%) with a similar inference speed. FMViT-L matches the performance of EfficientFormer-L7 \cite{DBLP:conf/nips/LiYWHETWR22} when compared to the hybrid approach, but the inference is 30\% and 96\% faster on TensorRT and CoreML, respectively. FMViT-S achieves 2.6\% higher performance (78.5\% vs. 75.9\%) with a comparable CoreML inference speed when compared to MobileOne-S1 \cite{DBLP:journals/corr/abs-2206-04040}, and CoreML achieves 11\% faster inference performance while achieving 2.9\% higher accuracy (78.5\% vs. 75.6\%). These results suggest that the proposed FMViT design is a feasible and promising paradigm.

\subsection{Object Detection and Instance Segmentation}
\subsubsection{Implementation}
We evaluate FMViT on object detection and instance segmentation tasks based on the Mask R-CNN \cite{DBLP:conf/iccv/HeGDG17} architecture and COCO2017 \cite{DBLP:conf/eccv/LinMBHPRDZ14}. Specifically, all our models are initially trained on ImageNet-1K and subsequently fine-tuned using the settings from previous work \cite{DBLP:conf/iccv/LiuL00W0LG21}. The AdamW optimizer is employed with a weight decay of 0.05, and the training spans 12 epochs. A warm-up of 500 iterations is performed during training, and the learning rate is decreased by 10 at the 8th and 11th epochs. Input resolution is 1344x800. For an equitable comparison, we solely assess backbone latency, and the testing environment aligns with that of classification.

\subsubsection{Comparison with State-of-the-art Models}
Table \ref{tab:table2} presents the evaluation results utilizing the Mask R-CNN architecture. For fairness, we exclusively measured the backbone latency. As per Table \ref{tab:table2}, FMViT-B surpasses ResNet101 3.7 $AP^b$ while achieving a 16\% faster inference on TensorRT. FMViT-B matches the inference speed of PoolFormer-S12 \cite{DBLP:conf/cvpr/YuLZSZWFY22}, on TensorRT but with a 6.8 $AP^b$ enhancement. Compared to EfficientFormer-L3, FMViT-B exhibits a 7\% faster inference on TensorRT and a 2.7 $AP^b$  superior performance. Against Next-ViT-S \cite{DBLP:journals/corr/abs-2207-05501}, FMViT-B demonstrates a 3.9 times faster inference on CoreML and a 0.3 $AP^b$ increased performance. FMViT-L outperforms EfficientFormer-L7 by 3.8 $AP^b$, and its inference is 32\% quicker on TensorRT. FMViT-L and ResNeSt101 have identical inference speeds on TensorRT, but FMViT-L shows a 1.2 $AP^b$ higher performance. The AP for masks exhibits a similar advantage. In conclusion, FMViT excels in object detection and instance segmentation while maintaining a reduced inference latency.

% Table generated by Excel2LaTeX from sheet 'det'
\begin{table*}[htbp]
  \centering
  \caption{Comparison of different backbones on Mask R-CNN-based object detection and instance segmentation tasks. The superscripts b and m denote box detection and mask instance segmentation.}
    \scalebox{0.6}{
    \begin{tabular}{l|ccc|ccc|cccccc}
    \hline
    \multicolumn{1}{c|}{\multirow{2}[4]{*}{backbone}} & Image & Param & FLOPs &       & Latency(ms) &       & \multicolumn{6}{c}{Mask R-CNN} \bigstrut\\
\cline{5-13}          & Size  & (M)   & (G)   & pytorch & TRT   & CoreML & $AP^b$   & $AP^b_{50}$ & $AP^b_{75}$ & $AP^m$   & $AP^m_{50}$ & $AP^m_{75}$ \bigstrut\\
    \hline
    ResNet101 & 1344x800 & 44.5  & 167.8  & 34.0  & 26.9  & 66.2  & 40.4  & 61.1  & 44.2  & 36.4  & 57.7  & 38.8  \bigstrut[t]\\
    ResNeXt101-32x4d & 1344x800 & 44.2  & 171.7  & 40.5  & 35.1  & 69.2  & 41.9  & 62.5  & 45.9  & 37.5  & 59.4  & 40.2  \\
    ResNeSt50 & 1344x800 & 27.5  & 115.6  & 35.9  & 24.1  & /     & 42.6  & /     & /     & 38.1  & /     & / \\
    Swin-T & 1344x800 & 47.8  & 264.0  & /     & /     & /     & 42.2  & 64.4  & 46.2  & 39.1  & 64.6  & 42.0  \\
    PoolFormer-S12 & 1344x800 & 11.9  & 39.0  & 48.3  & 23.1  & 32.1  & 37.3  & 59.0  & 40.1  & 34.6  & 55.8  & 36.9  \\
    PoolFormer-S24 & 1344x800 & 21.4  & 73.1  & 95.8  & 45.6  & 58.8  & 40.1  & 62.2  & 43.4  & 37.0  & 59.1  & 39.6  \\
    EfficientFormer-L3 & 1344x800 & 31.4  & 89.6  & 30.5  & 24.7  & 66.1  & 41.4  & 63.9  & 44.7  & 38.1  & 61.0  & 40.4  \\
    Next-ViT-S & 1344x800 & 51.8  & 301.1  & 45.5  & 34.7  & 843.0  & 43.8  & 65.7  & 47.9  & 39.8  & 63.0  & 42.6  \\
    \textbf{FMViT-B} & \textbf{1344x800} & \textbf{72.1 } & \textbf{269.9 } & \textbf{27.8 } & \textbf{23.1 } & \textbf{213.8 } & \textbf{44.1 } & \textbf{66.1 } & \textbf{48.0 } & \textbf{41.8 } & \textbf{65.1 } & \textbf{45.0 } \bigstrut[b]\\
    \hline
    ResNeXt101-64x4d & 1344x800 & 83.5  & 332.5  & 62.8  & 65.2  & 107.8  & 42.8  & 63.8  & 47.3  & 38.4  & 60.6  & 41.3  \bigstrut[t]\\
    ResNeSt101 & 1344x800 & 48.3  & 219.4  & 63.2  & 42.1  & /     & 45.2  & /     & /     & 40.2  & /     & / \\
    Swin-S & 1344x800 & 69.1  & 354.0  & /     & /     & /     & 44.8  & 66.6  & 48.9  & 40.9  & 63.4  & 44.2  \\
    PoolFormer-S36 & 1344x800 & 30.9  & 107.2  & 143.3  & 68.9  & 89.2  & 41.0  & 63.1  & 44.8  & 37.7  & 60.1  & 40.0  \\
    EfficientFormer-L7 & 1344x800 & 82.2  & 228.6  & 66.9  & 55.6  & 155.5  & 42.6  & 65.1  & 46.1  & 39.0  & 62.2  & 41.7  \\
    Next-ViT-B & 1344x800 & 64.9  & 354.1  & 61.8  & 48.1  & 932.0  & 45.3  & 67.0  & 49.7  & 41.0  & 64.2  & 44.2  \\
    \textbf{FMViT-L} & \textbf{1344x800} & \textbf{94.9 } & \textbf{333.0 } & \textbf{45.4 } & \textbf{42.0 } & \textbf{276.0 } & \textbf{46.4 } & \textbf{68.1 } & \textbf{51.2 } & \textbf{41.9 } & \textbf{65.4 } & \textbf{45.1 } \bigstrut[b]\\
    \hline
    \end{tabular}% 
    }
  \label{tab:table2}%
\end{table*}%

\subsection{ADE20K Semantic Segmentation}
\subsubsection{Implementation}
We conducted semantic segmentation tests utilizing the ADE20K \cite{DBLP:conf/cvpr/ZhouZPFB017} dataset, which comprises approximately 20K training images and 2K validation images across 150 categories. For a fair comparison, we adhered to the training protocol of the preceding vision transformer on the Semantic FPN \cite{DBLP:conf/cvpr/KirillovGHD19} framework. The model was initially pre-trained on ImageNet-1K at a resolution of 224x224, then trained on ADE20K with an input size of 512x512. For the Semantic FPN framework, we employed the AdamW optimizer with a learning rate and weight decay of 0.0001. We trained the entire network for 40K iterations with a total batch size of 32. Given the complexity of implementing various modules of Mask R-CNN on TensorRT and CoreML, we limited our latency assessment to the backbone for a fair comparison, maintaining the same test setup as for classification. For simplicity, We used an input size of 512x512 to measure latency.

\subsubsection{Comparison with State-of-the-art Models}
Table \ref{tab:table3} illustrates that FMViT-B surpasses ResNet101 \cite{DBLP:conf/cvpr/HeZRS16} by 4.7 mIoU while maintaining consistent inference speed on TensorRT and CoreML. It exceeds Swin-T\cite{DBLP:conf/iccv/LiuL00W0LG21} by 2.0 mIoU. Compared to PoolFormer-S24 \cite{DBLP:conf/cvpr/YuLZSZWFY22}, it achieves 3.2 mIoU higher performance and is 8\% faster in TensorRT inference. Our performance improvement of 0.5 mIoU is accompanied by 18\% and 43\% faster inference on TensorRT and CoreML, respectively, compared to Next-ViT-S. FMViT-L outperforms Swin-S by 1.7 mIoU and is 4.5 mIoU higher than CoreML while being 25 times faster than ResNeSt101 \cite{DBLP:conf/cvpr/0005WZZLZSHMMLS22}. It matches the inference performance of PoolFormer-S36 but with a 4.9 mIoU advantage. Inference on TensorRT and CoreML is 2.5\% and 29\% faster than Next-ViT-B, with comparable mIoU. Comprehensive testing indicates that our FMViT holds significant potential for segmentation tasks.

% Table generated by Excel2LaTeX from sheet 'seg'
\begin{table}[htbp]
  \centering
  \caption{Comparison of different backbones on the ADE20K semantic segmentation task.}
  \scalebox{0.5}{
    \begin{tabular}{l|c|ccc|ccc}
    \hline
    \multicolumn{1}{c|}{\multirow{2}[2]{*}{backbone}} & Image &       & Latency(ms) &       & \multicolumn{3}{c}{Semantic FPN 80k} \bigstrut[t]\\
          & Size  & pytorch & TRT   & CoreML & Param(M) & FLOPs(G) & mIoU \bigstrut[b]\\
    \hline
    ResNet101 & 512x512 & 12.9  & 7.4   & 10.9  & 44.5  & 40.9  & 38.8  \bigstrut[t]\\
    ResNeXt101-32x4d & 512x512 & 13.8  & 10.1  & 14.1  & 44.2  & 41.8  & 39.7  \\
    ResNeSt50 & 512x512 & 12.7  & 6.5   & 356.0  & 27.5  & 28.2  & 39.7  \\
    Swin-T & 512x512 & /     & /     & /     & 31.9  & 182.0  & 41.5  \\
    PoolFormer-S12 & 512x512 & 9.1   & 4.1   & 6.2   & 11.9  & 9.5   & 37.2  \\
    PoolFormer-S24 & 512x512 & 17.8  & 7.8   & 10.8  & 21.4  & 17.8  & 40.3  \\
    EfficientFormer-L3 & 512x512 & 10.5  & 6.6   & 10.0  & 31.4  & 20.7  & 43.5  \\
    Next-ViT-S & 512x512 & 14.5  & 8.5   & 15.6  & 36.3  & 52.0  & 43.0  \\
    \textbf{FMViT-B} & \textbf{512x512} & \textbf{9.0 } & \textbf{7.2 } & \textbf{10.9 } & \textbf{24.8 } & \textbf{22.2 } & \textbf{43.5 } \bigstrut[b]\\
    \hline
    ResNeXt101-64x4d & 512x512 & 17.9  & 17.4  & 22.5  & 83.5  & 81.1  & 40.2  \bigstrut[t]\\
    ResNeSt101 & 512x512 & 25.1  & 11.3  & 423.0  & 48.3  & 53.5  & 42.4  \\
    Swin-S & 512x512 & /     & /     & /     & 53.2  & 274.0  & 45.2  \\
    PoolFormer-S36 & 512x512 & 26.4  & 11.5  & 15.2  & 30.9  & 26.1  & 42.0  \\
    PoolFormer-M36 & 512x512 & 41.5  & 16.5  & 22.1  & 56.1  & 46.0  & 42.4  \\
    EfficientFormer-L7 & 512x512 & 16.9  & 14.4  & 21.4  & 82.2  & 53.6  & 45.1  \\
    Next-ViT-B & 512x512 & 20.8  & 12.2  & 21.1  & 49.3  & 64.9  & 47.1  \\
    \textbf{FMViT-L} & \textbf{512x512} & \textbf{15.2 } & \textbf{11.9 } & \textbf{16.3 } & \textbf{36.5 } & \textbf{37.4 } & \textbf{46.9 } \bigstrut[b]\\
    \hline
    \end{tabular}%
    }
  \label{tab:table3}%
\end{table}%

\subsection{Ablation Study}
We established a series of experiments to validate the efficiency of the FMB, gMPL, and RLMHSA modules within FMViT, as depicted in Table \ref{tab:table4}. Here, we incorporated our proposed modules into the FMViT-T0 model and adhered to the same training methodology as the original model. RLMHSA substitutes the traditional MHSA, gMPL supersedes the standard MPL, and FMB is not utilized for mixing; only the standard MHSA output features are directly fed into the MLP.

The experimental findings indicate that substituting the standard MHSA with our more streamlined RLMHSA decreases classification performance despite increasing parameters and FLOPs. When the conventional MLP module is replaced with the convolutional multi-group reparameterized gMLP, the number of parameters and FLOPs during the inference stage remain comparable, yet classification performance improves. Lastly, introducing the FMB module significantly increases the number of parameters and FLOPs, but it also boosts classification accuracy to the final level.

% Table generated by Excel2LaTeX from sheet 'Sheet4 (3)'
\begin{table}[htbp]
  \centering
  \caption{Compare different modules.}
  \scalebox{0.8}{
    \begin{tabular}{c|c|c|c|c|c}
    \hline
    FMB   & gMPL  & RLMHSA & Param(M) & FLOPs(M) & Top-1(\%) \bigstrut\\
    \hline
          &       &       & 2.61  & 479.24  & 68.63  \bigstrut[t]\\
          &       & \checkmark     & 2.25  & 338.05  & 67.34  \\
          & \checkmark     & \checkmark     & 2.25  & 338.05  & 69.30  \\
    \checkmark     & \checkmark     & \checkmark     & 2.52  & 395.30  & 72.90  \bigstrut[b]\\
    \hline
    \end{tabular}%
    }
  \label{tab:table4}%
\end{table}%

\subsection{Visualization}
\begin{figure}[h]
    \centering
    \includegraphics[width=0.4\textwidth]{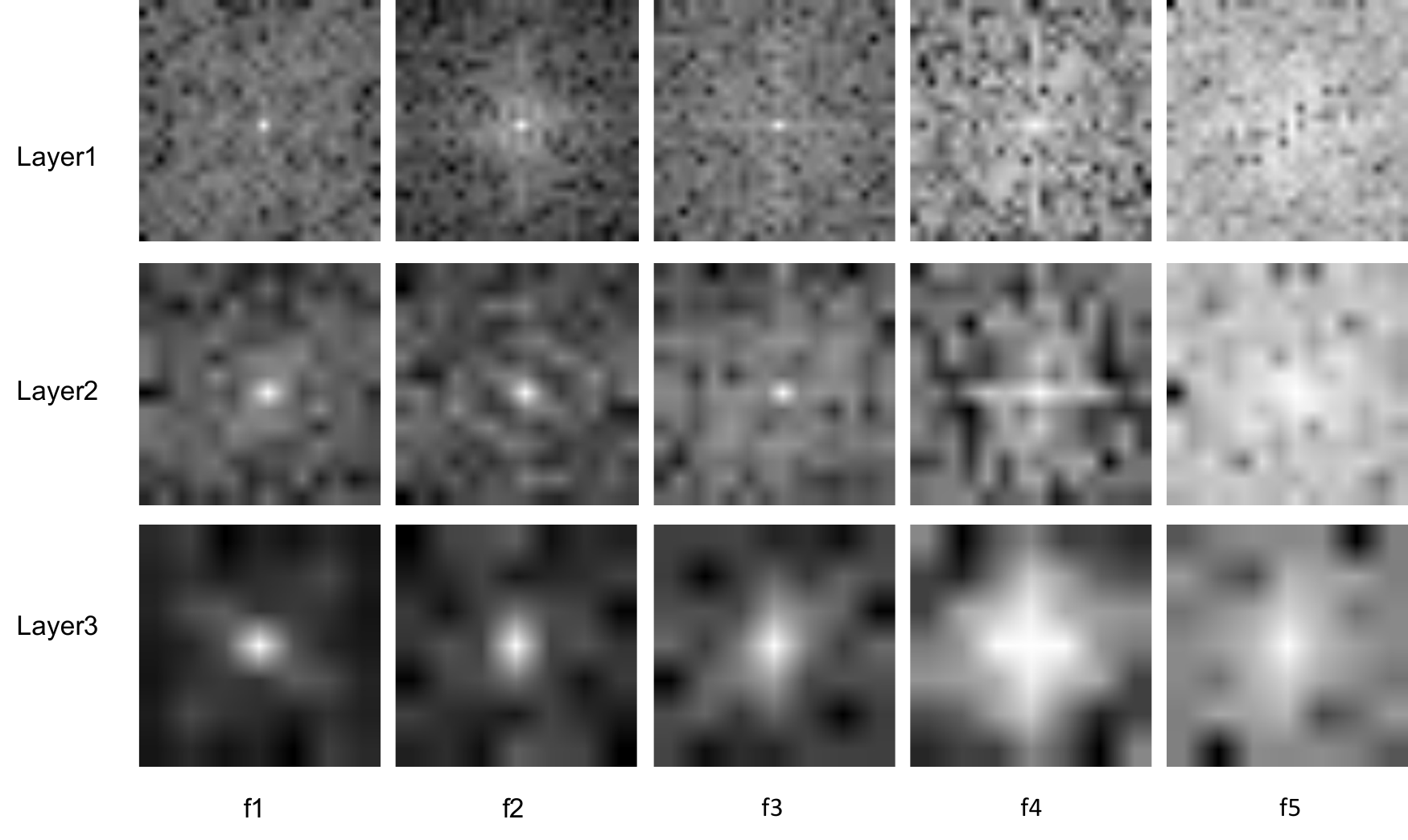}
    \caption{The Fourier spectrum of the output features of different modules in FMViT.}
    \label{fig:fig3}
\end{figure}

NextViT \cite{DBLP:journals/corr/abs-2207-05501} establishes that CNN convolutional blocks favor high-frequency signals, while ViT is inclined towards low-frequency signals. Our proposed FMB simultaneously captures diverse high-frequency and low-frequency signals, thereby enabling the acquisition of richer texture information and more precise global information, enhancing the modeling capability of FMViT. To better understand FMViT, we visualize the Fourier spectrum of RLMHSA output features in FMVIT-T0 at both high and low frequencies. Within RLMHSA are five features with varying frequencies, each representing different frequency characteristics, denoted as f1–f5. Figure \ref{fig:fig3} illustrates this. The RLMHSA output feature, f1, captures the low-frequency signal, suggesting that RLMHSA excels at modeling global information. f2-f5, the outputs of various CFBs, capture different high-frequency signals. Each output corresponds to a distinct frequency, so they are proficient at handling various textures. The fusion of f1-f5 frequency features enhances the model's expressive capacity.

\section{Conclusion}
In this study, we introduce a hybrid ViT architecture optimized for efficient deployment on mobile devices and server GPUs. This architecture enhances the model's predictive power by amalgamating high-frequency and low-frequency features at varying frequencies, thereby bolstering the model's capacity to capture both local and global information. Experimental results demonstrate that FMViT achieves state-of-the-art latency and accuracy trade-offs across multiple vision tasks, including image classification, object detection, and semantic segmentation. However, the models we provided are stacked together without further scrutiny. Future work could employ Network Architecture Search (NAS) or other stacking methods to explore the impact of different combination models on performance.

\bibliography{aaai24}

\end{document}